\pdfoutput=1

\documentclass[11pt]{article}

\usepackage{authblk}

\usepackage{acl}

\usepackage{times}
\usepackage{latexsym}

\usepackage[T1]{fontenc}

\usepackage[utf8]{inputenc}

\usepackage{microtype}

\usepackage{graphicx}
\usepackage{multirow}
\usepackage{booktabs}
\usepackage{enumitem}
\usepackage{changepage}

\title{Classical Out-of-Distribution Detection Methods Benchmark \\in Text Classification Tasks}

\author[1,2]{\textbf{Mateusz Baran}}
\author[1]{\textbf{Joanna Baran}}
\author[1,2]{\textbf{Mateusz Wójcik}}
\author[1,3]{\textbf{Maciej Zięba}}
\author[2]{\textbf{Adam Gonczarek}}
\affil[1]{Wroclaw University of Science and Technology \protect\\ \texttt{\{firstname.lastname\}@pwr.edu.pl}}
\affil[2]{Alphamoon Ltd., Wrocław \protect\\ \texttt{\{firstname.lastname\}@alphamoon.ai}}
\affil[3]{Tooploox Ltd., Wrocław}

\begin{document}
\maketitle
\begin{abstract}


State-of-the-art models can perform well in controlled environments, but they often struggle when presented with out-of-distribution (OOD) examples, making OOD detection a critical component of NLP systems.
In this paper, we focus on highlighting the limitations of existing approaches to OOD detection in NLP.
Specifically, we evaluated eight OOD detection methods that are easily integrable into existing NLP systems and require no additional OOD data or model modifications.
One of our contributions is providing a well-structured research environment that allows for full reproducibility of the results. 
Additionally, our analysis shows that existing OOD detection methods for NLP tasks are not yet sufficiently sensitive to capture all samples characterized by various types of distributional shifts.
Particularly challenging testing scenarios arise in cases of background shift and randomly shuffled word order within in domain texts. 
This highlights the need for future work to develop more effective OOD detection approaches for the NLP problems, and our work provides a well-defined foundation for further research in this area.

\end{abstract}

\section{Introduction}

Systems based on artificial intelligence (AI) have to be safe and trustworthy~\cite{DBLP:journals/corr/AmodeiOSCSM16}.
Ensuring user reliance on these systems requires a cautious approach in making predictions. 
AI tools should avoid decisions on examples that significantly deviate from the training data. 
This is especially risky when the classifier shows excessive confidence in its incorrect decisions, leading to the propagation of errors in the system pipeline~\cite{doi/10.2759/346720}.
However, current models are often trained under the closed-world assumption, limited to specific domains~\cite{parketal2022efficient}. Test sets drawn from the same domain for evaluation may not reflect real-world scenarios accurately~\cite{10.5555/3495724.3495759}. This poses challenges when deploying such models in production environments~\cite{NEURIPS2022_7a969c30}.


\begin{figure}[ht]
    \centering
    \includegraphics[width=1.0\linewidth]{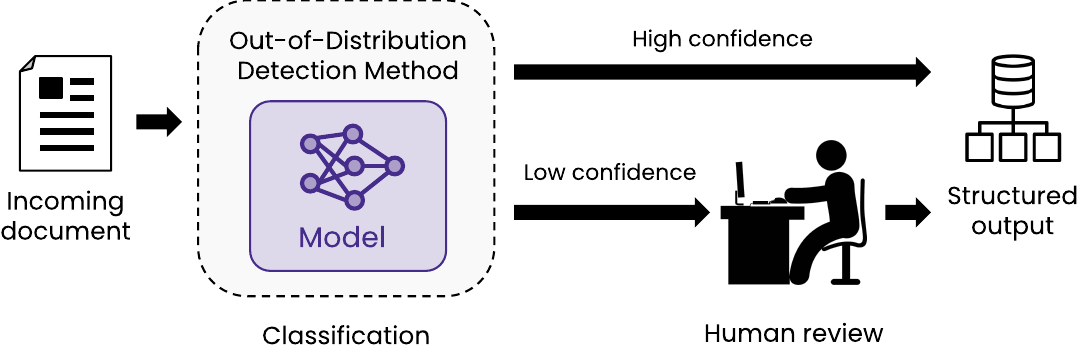}
    \caption{Trustworthy mechanism in document processing platform. Classification models need additional method to detect OOD samples and provide them to human review.}
    \label{fig:hil}
\end{figure}

Real-world data is often completely different from training one. The change in data distribution can be caused by several factors such as user behavior, legal regulations, market trends or seasonal changes.
In an \textit{open-world} scenario, the AI-based system can be even exposed to inputs that deviate from the trained task. A significant risk that may arise is the possibility of model overconfidence while predicting data of this nature.
As a~result, there is a business need for detecting examples outside the domain~\cite{hendrycks17baseline}. Out-of-distribution (OOD) detection techniques can be well applied in a production system with human-in-the-loop technology~\cite{WU2022364}, where it is important to quickly identify whether an input sample is characterized by a distributional shift.
Such an example should be handled then by a human expert in order to avoid potential misclassification by the model.
The essence of such systems is to find a trade-off between the accuracy and automation~\cite{10.1007/s10462-022-10246-w} (Figure~\ref{fig:hil}).
This way, the model can achieve the highest possible performance on in-distribution (ID) data and difficult shifted data can be given to human verification, thus increasing the credibility of the overall system.
The bottleneck here is a well-designed OOD detection method, which must be sensitive enough to capture all examples outside the domain.

The problem of OOD identification is mainly investigated for vision classification tasks~\cite{yang2022openood, kuan2022basics}, whereas in the field of NLP, studies on this topic are limited.
We fill the missing gap by proposing a comprehensive analysis of existing OOD approaches for text classification tasks. 
In this work, we focus on the \textbf{post-hoc} techniques which are most suitable for business applications i.e. they have to fulfil the requirement of smooth integration into existing systems, without the need for additional OOD training data or any changes in model architecture. 
Ultimately, we evaluated eight methods in two different scenarios.
The first one includes grouping test data into three splits according to the similarity to the in-distribution set: \textit{Near-OOD}, \textit{Far-OOD} and \textit{Distinct-OOD}~\cite{yang2021oodsurvey}. 
The AI system is evaluated based on the degree of domain difference between training and test samples.
The second scenario considers the division of datasets according to the shift of distribution~\cite{arora2021types}.
There are many categories of distribution shift~\cite{DBLP:journals/corr/abs-2210-03050}, but in this study, we consider two types -- semantic and background.
\textbf{Semantic shift} occurs when new labels appear, which may be due to the lack of a sufficient number of classes representing the training data or the emergence of new classes over time.
In distinction, the \textbf{background shift} is class independent. 
It appears when the characteristic features of text change (e.g. source origin, writing style), which can happen even within the same class. The reason may be language evolution, regional conditions, etc. -- such factors are difficult to predict and adequately address in the training set.
By preparing data separated into different kinds of shift, we gain an in-depth insight into the origin of the data, on which a particular OOD detection method performs better or worse.

We also provide a well-structured research environment that allows the full reproducibility of the achieved outcomes and evaluation of another NLP models.
The source code is available on GitHub\footnote{  
\url{https://github.com/mateuszbaransanok/TrustworthyAI}
}. 
To summarize, our contribution is as follows:
\begin{itemize}[noitemsep,nolistsep]
\item we adjust the existing OOD detection techniques to the text classification problems,
\item we comprehensively evaluate the revised methods using two different scenarios tailored to the NLP domain,
\item we deliver the complete experimental framework for evaluating the OOD methods.
\end{itemize}

\section{Related Work}
In recent years, there has been a growing interest in developing robust methods that can detect out-of-distribution examples.
The work of~\citet{hendrycks17baseline} has played a significant role in advancing this field. Their Maximum Softmax Probability (MSP) method, which relies on the softmax output of a neural network, has become a reference for subsequent research and still remains as the solid baseline approach~\cite{zhang2023unsupervised}.
The benefit of the MSP was its independence from the specific task domain.
Since then, many researchers have extended this method or proposed novel techniques to address the challenge of detecting OOD~data.

The first to popularize the interest in the OOD topic were computer vision (CV) researchers~\cite{pmlr-v15-bengio11b}.
The emerged techniques in this field were summarized in a survey by~\citet{yang2021oodsurvey}. 
The authors proposed a unified framework that groups OOD detection methods into categories based on their common underlying mechanisms. 
Among them, the following ones can be distinguished:
(1)~\textbf{output-based}~\cite{liu2021energybased, liang2018enhancing} techniques which detect OOD samples based on output vector obtained by classification model for given input;
(2)~\textbf{gradient-based}~\cite{huang2021importance} focus on analyzing the fluctuation of the gradient flow through the model layers to verify that the input is OOD;
(3)~\textbf{density-based}~\cite{ZongSMCLCC18} methods involve modeling a density function from the training set and then determining whether a new example belongs to the same distribution;
(4)~\textbf{distance-based}~\cite{sun2022outofdistribution, DBLP:journals/corr/abs-2106-09022} measure the dissimilarity between a~new input and the training data by computing standard metrics such as cosine similarity, Euclidean or Mahalanobis distance.
Another work of~\citet{yang2022openood} provides a comprehensive evaluation of 13 methods for OOD detection in CV. 
Notably, the experimental results show that simple preprocessing techniques can be highly effective, outperforming even more sophisticated methods in identifying OOD examples. 
In addition, post-hoc methods have demonstrated considerable effectiveness in OOD detection and have made significant impact in this task.
The NLP community is also more and more interested in addressing the challenge of OOD detection data, especially after the appearance of text processing automation systems. 
Despite the expectation that pre-trained language models (PLMs) would generalize well to unseen data, many existing transformer-based architectures perform poorly in an open-world assumption setup. 
This was proven by the work~\cite{yang2022gluex} where the authors created the GLUE-X benchmark to reliably test the robustness of PLMs against OOD samples exposure, without using any of the previously mentioned techniques dedicated to OOD.
Their achieved results confirm the necessity of further development of OOD detection methods. 
Currently, researchers are continuously proposing techniques tailored for the NLP tasks~\cite{rawat2021pnpood, zhouetal2021contrastive}, revisiting existing ones~\cite{Podolskiy2021RevisitingMD} or designing completely novel approaches that can address specific shifts in data distribution~\cite{arora2021types, Chen2023FineTuningDG}.
The latter two publications particularly highlight the importance of dividing datasets into semantic and background shift sets, as they provide valuable findings and a better understanding of how the model works on different data types.

Evidently, there have been several NLP articles addressing OOD detection, but their comparison to existing methods has been limited. A comprehensive study which evaluates various OOD detection approaches on a larger scale and addressing the specific needs of businesses is still lacking.
To fill this gap, we have developed a benchmark that provides a fair comparison of these techniques while testing their performance across different distributional shift scenarios. All the selected methods have been inspired by CV achievements, and we have specifically chosen those that can be easily integrated into an existing AI system with minimal complexity.


\section{Benchmark Outline}
This section provides an overview of the datasets and the model architecture, with a detailed description of the techniques reimplemented in our benchmark for detecting out-of-domain examples. 
The metrics used for evaluating the effectiveness of the detection methods  are also presented.

\subsection{Datasets}
\label{sec:datasets}

\noindent\textbf{News Category Dataset}~\cite{misra2022news} is one of the biggest news dataset.
It contains around 210k news headlines from HuffPost published between 2012 and 2022.
The dataset comprises of 42 classes that are heavily imbalanced. 
Therefore, the most similar classes were combined to avoid confusion between similar classes. 
Ultimately, we obtained 17 representative classes.

\noindent\textbf{Twitter Topic Classification}~\cite{dimosthenis-etal-2022-twitter} is a topic classification dataset collected from Twitter posts. 
It consists of 3184 high-quality tweets that have been assigned to one of six classes.

\noindent\textbf{SST-2} (The Stanford Sentiment Treebank)~\cite{socher} is a corpus with fully labeled parse trees that allows for an analysis of the compositional effects in language sentiment. 
The corpus includes almost 70k sentences extracted from movie reviews. 
Sentences were annotated with regard to their polarization (positive or negative).

\noindent\textbf{IMDB}~\cite{maas-EtAl:2011:ACL-HLT2011} is a large collection of movie reviews from the Internet Movie Database created for the binary sentiment classification task.
According to the original 10-point movie rating scale from the website, the dataset samples were filtered to include only highly polarized texts annotated as positive ($\ge7$) or negative ($\le4$).

\noindent\textbf{Yelp Polarity Review}~\cite{zhangCharacterlevelConvolutionalNetworks2015} dataset includes almost 600k customer reviews which are labeled as positive or negative based on the number of stars given by the reviewer. 
Specifically, texts with $\le2$ stars are labeled as negative, while those with $\ge3$ are labeled as positive.
Due to the large size of the dataset, we created a smaller version by randomly selecting a subset of 75k reviews.

\noindent\textbf{Language Detection Dataset}~\cite{LanguageDataset} is a~small dataset for language detection task. 
It contains texts in 17 different languages.
For benchmark purposes, we filter out languages that do not use Latin alphabet.
We've also excluded English texts to create a clear out-of-distribution dataset. 
Finally, dataset consist around 6k samples and all of them are used for OOD evaluation.

\noindent\textbf{20 Newsgroups}~\cite{Newsgroups20} consists of around 18k newsgroups posts on 20 topics. 
It is divided in two sets for training and evaluation. 
Moreover, we allocated an additional subset from the training set for validation purposes.

\subsection{Model}
In all experiments, we used transformer-based~\cite{NIPS2017_3f5ee243} RoBERTa\textsubscript{base}~\cite{DBLP:journals/corr/abs-1907-11692} model as a backbone with a fully connected layer as a classification head. 
The model was pretrained on English corpora, but it supports multiple languages.

\subsection{Methods}
We decided to compare \textbf{post-hoc} methods that are suitable to apply to trained models.
They mainly use information based on model statistics such as intermediate layer values, gradients or non-deterministic properties of dropout regularization, etc.
Their implementation is technically straightforward and independent of the type of model used.

\begin{figure}[ht]
    \centering
    \includegraphics[width=0.7\linewidth]{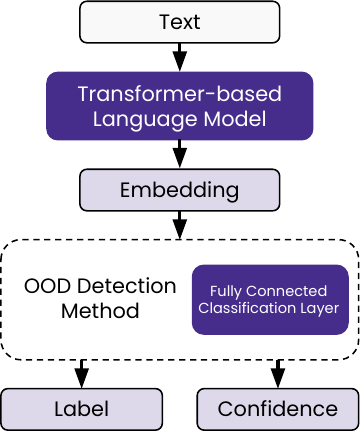}
    \caption{Benchmark schema -- fine-tuned PLM-based classifier followed by OOD detection method.}
    \label{fig:schema}
\end{figure}

An overview of our benchmark methodology is outlined in Figure~\ref{fig:schema}.
In addition to label prediction, we obtain a real-valued \textit{confidence} score that indicates the level of confidence that the model has in whether the given sample belongs to the ID data.
We reimplemented eight OOD detection techniques and adapted them to the NLP classification pipeline.
\begin{enumerate}[label=(\arabic*),wide,labelindent=0pt,itemsep=1pt]
    \item \textbf{Maximum Softmax Probability (MSP)}~\cite{hendrycks17baseline} employs the softmax score to check the certainty of whether an example belongs to a domain -- we refer to it as the baseline method in our work.
    \item \textbf{Energy-based}~\cite{liu2021energybased} uses an energy score function to indicate model confidence.
    \item \textbf{Rectified Activations (ReAct)}~\cite{sun2021react} is a simple technique for reducing model overconfidence on OOD examples by truncating the high activations during evaluation.
    \item \textbf{KL-Matching (KLM)}~\cite{hendrycks2022scaling} calculates the minimum KL-divergence between the softmax probabilities and the mean class-conditional distributions.
    \item \textbf{GradNorm}~\cite{huang2021importance} utilizes information obtained from the gradient space of model's classification layer. 
    This approach uses the vector norm of gradients to distinguish between ID and OOD samples, with the assumption that higher norm values correspond to in-distribution data.
    \item  \textbf{Directed Sparisification (DICE)}~\cite{sun2022dice} selectively chooses a subset of weights through sparsification, which helps to eliminate irrelevant information from the output.
    \item \textbf{Virtual-logit Matching (ViM)}~\cite{wang2022vim} combines information from feature space (PLM embedding) and output logits, providing both class-agnostic and class-dependent knowledge simultaneously for better separation of OOD data.
    \item \textbf{K-nearest neighbors (KNN)}~\cite{sun2022outofdistribution} computes the distance between the embedding of an input example and the embeddings of the training set, and uses it to determine whether the example belongs to the ID or not.
\end{enumerate}

The first four methods use signals originating from the output layer of the model. 
GradNorm focuses solely on the gradients that flow through the classification head, while methods from 6 to 8 operate on the embedding of a PLM.
Most techniques (specifically no. 3-4, 6-8) need an initial configuration on the training or validation set to estimate the required statistics for ID data.
To ensure consistency in the benchmarking process, the hyperparameters for the above methods were set to the values recommended in their original papers.

\subsection{Metrics}
To compare the chosen methods, we used three the most common metrics for OOD detection.

\noindent\textbf{AUROC} calculates the area under the Receiver Operating Characteristic (ROC) curve.
The ROC curve plots the true positive rate against the false positive rate, and a larger area under the curve indicates better performance.
This was used as our primary evaluation metric.

\noindent\textbf{AUPR-IN} measures the area under the Precision-Recall (PR) curve.
The PR curve displays how well the method can identify true positives with high precision, and AUPR provides a measure of overall performance.
The \textit{"IN"} suffix indicates that this metric pertains to in-distribution data.

\noindent\textbf{FPR@95} is the false positive rate when the true positive rate is set to 95\%.
Lower scores indicate better performance.

\begin{table}[ht]
    \centering
    \caption{Datasets setup for experiments.}
    \label{tab:datasets}
    
\scalebox{0.74}{
\begin{tabular}{lccc}
\toprule
 \textbf{Dataset} &  \textbf{\#Classes} &    \textbf{Train / Val / Test} & \textbf{Avg. words} \\
\midrule
    NC/I &          7 & 66223 / 26475 / 39688  & 9.95 \\
    NC/O &         10 &         - / - / 48522  & 9.77 \\
 Twitter &          6 &          - / - / 3184  & 29.80 \\
    IMDB &          2 &  25000 / 5000 / 20000  & 231.15 \\
   SST-2 &          2 &  43221 / 5000 / 20000  & 9.53 \\
    Yelp &          2 &  50000 / 5000 / 20000  & 133.11 \\
Language &          9 &          - / - / 5864  & 19.08 \\
   NCR/I &          7 &         - / - / 39688  & 9.95 \\
   NCR/O &         10 &         - / - / 48522  & 9.77 \\
Computer &          5 &     2965 / 456 / 1460  & 218.63 \\
Politics &          4 &      1959 / 315 / 979  & 406.53 \\
  Sports &          4 &     2363 / 432 / 1182  & 224.43 \\
\bottomrule
\end{tabular}
}
\end{table}

\section{Data Preparation}
\label{sec:data_preparation}
In our study, we have paid particular attention to provide a complete and unbiased comparison of OOD detection methods. 
To achieve this goal, we adopted two diverse perspectives: one inspired by the field of computer vision~\cite{yang2022openood} and the other drawn from works dedicated to the NLP domain~\cite{rawat2021pnpood, arora2021types}.

\subsection{Scenario 1}
\label{subsec:data_scenario_1}
The first perspective intends to provide a detailed analysis of considered techniques based on the similarity between OOD examples and the training set. 
The degree of similarity is defined here in a~human-intuitive way, taking into account such factors as thematic proximity, task dissimilarity or the sentence correctness.

As a base in-distribution data, we chose \textit{News Category} dataset using the seven most popular classes (\textbf{NC/I}).
The remaining classes were considered as out-of-distribution split (\textbf{NC/O}) which represents data in close semantic shift.
The \textit{Twitter Topic Classification} dataset has categories that are similar to those in the \textit{News Category} dataset, but the sentence construction is significantly different.
Both sets create the \textbf{Near-OOD} data setup.
Another prepared collection, \textbf{Far-OOD} includes datasets with reviews of movies, hotels and restaurants that are vastly different from \textit{NC/I} data -- it is a connection of \textit{SST-2}, \textit{Yelp} and \textit{IMDB}.
Additionally, we prepared one more group named \textbf{Distinct-OOD} 
containing \textit{Language Detection} dataset. 
With the inclusion of non-English texts there, we obtain a~distinct set of tokens that the RoBERTa model has not encountered before, creating a completely separate dataset from the in-distribution data.

Finally, we also designed two collections derived from the \textit{News Category} dataset by randomly shuffling words from all those available within each category. 
The new dataset, called \textit{News Category Random}, retained the original number of examples and the number of words in each sample. 
These sets aimed to examine the classification system behavior when presented with input sentences that are completely disrupted from their original context. 
The previous partition into ID (\textbf{NCR/I}) and OOD (\textbf{NCR/O}) subsets was maintained. 

\subsection{Scenario 2}
\label{subsec:data_scenario_2}
This scenario investigated the performance of detection methods for OOD examples under semantic and background shift.
For semantic shift, we utilized the \textit{20 Newsgroups} dataset that is a hierarchical collection of documents. 
Among the four top-level categories, we selected three - \textbf{Computer}, \textbf{Sports}, and \textbf{Politics} - as training sets for the model, while excluding the \textit{"misc"} category due to potential data leakage issues.
Subsequently, we generated various combinations of these categories, treating each one in turn as an in-distribution set, while considering the others as a OOD data. 
For example, the model could be trained on the samples from  Computer class (ID dataset) and evaluated later on Sports and Politics (OOD).

In order to test the impact of background shift, we took three sentiment classification datasets -- \textit{IMDB}, \textit{SST-2} and \textit{Yelp}, which are based on user reviews and represent different domains. Although these datasets have similar linguistic properties, the topics they address are distinct. Again, we constructed various combinations of these collections by treating each one as the ID set and the others as OOD sets.

\begin{table*}[ht]
    \centering
    \caption{AUROC (\%) and standard deviations for methods evaluated on datasets from first scenario.}
    \label{tab:scenario1}
    \begin{tabular}{lllllllll}
\toprule
  & \multicolumn{2}{c}{\textbf{Near-OOD}} & \multicolumn{3}{c}{\textbf{Far-OOD}} & \multicolumn{3}{c}{\textbf{Distinct-OOD}} \\
  \cmidrule(lr){2-3} \cmidrule(lr){4-6} \cmidrule(lr){7-9}
  \textbf{Method} &                                    \textbf{NC/O} &                                 \textbf{Twitter} &                                    \textbf{IMDB} &                                   \textbf{SST-2} &                                    \textbf{Yelp} &                                \textbf{Language} &                                   \textbf{NCR/I} &                                   \textbf{NCR/O} \\
\midrule
     MSP &          74.2\textsubscript{\textpm0.3} &          74.8\textsubscript{\textpm2.4} &          96.6\textsubscript{\textpm3.1} &          84.2\textsubscript{\textpm3.3} &          95.3\textsubscript{\textpm1.5} &          95.1\textsubscript{\textpm1.9} &          59.0\textsubscript{\textpm0.8} &          80.5\textsubscript{\textpm0.6} \\
  Energy &          77.6\textsubscript{\textpm0.4} &          84.8\textsubscript{\textpm1.9} &          99.6\textsubscript{\textpm0.5} &          92.6\textsubscript{\textpm2.6} &          98.6\textsubscript{\textpm0.7} &          98.7\textsubscript{\textpm0.6} &          60.1\textsubscript{\textpm1.0} &          84.9\textsubscript{\textpm0.7} \\
GradNorm &          77.2\textsubscript{\textpm0.5} &          81.8\textsubscript{\textpm2.7} &          99.0\textsubscript{\textpm1.1} &          90.8\textsubscript{\textpm2.2} &          97.8\textsubscript{\textpm0.8} &          97.8\textsubscript{\textpm0.7} &          60.5\textsubscript{\textpm1.4} &          85.0\textsubscript{\textpm0.8} \\
     KLM &          62.9\textsubscript{\textpm0.4} &          54.0\textsubscript{\textpm3.8} &          92.5\textsubscript{\textpm6.2} &          67.7\textsubscript{\textpm4.6} &          88.9\textsubscript{\textpm3.7} &          86.7\textsubscript{\textpm3.9} &          50.6\textsubscript{\textpm0.1} &          68.5\textsubscript{\textpm0.6} \\
   ReAct &          77.5\textsubscript{\textpm0.4} &          84.5\textsubscript{\textpm2.0} &          99.6\textsubscript{\textpm0.5} &          92.4\textsubscript{\textpm2.8} &          98.6\textsubscript{\textpm0.7} &          98.7\textsubscript{\textpm0.6} &          60.0\textsubscript{\textpm1.0} &          84.7\textsubscript{\textpm0.7} \\
    DICE &          58.2\textsubscript{\textpm0.6} &          60.9\textsubscript{\textpm3.2} &          76.6\textsubscript{\textpm5.8} &          60.9\textsubscript{\textpm1.4} &          84.4\textsubscript{\textpm2.2} &          69.3\textsubscript{\textpm2.8} &          51.2\textsubscript{\textpm0.9} &          60.4\textsubscript{\textpm1.4} \\
     KNN & \textbf{80.1\textsubscript{\textpm0.2}} & \textbf{92.9\textsubscript{\textpm1.2}} & \textbf{99.8\textsubscript{\textpm0.1}} & \textbf{96.4\textsubscript{\textpm1.1}} & \textbf{99.5\textsubscript{\textpm0.1}} & \textbf{99.6\textsubscript{\textpm0.1}} & \textbf{67.6\textsubscript{\textpm1.3}} & \textbf{88.7\textsubscript{\textpm0.5}} \\
     ViM &          79.9\textsubscript{\textpm0.2} &          89.2\textsubscript{\textpm1.5} &          90.6\textsubscript{\textpm3.1} &          96.0\textsubscript{\textpm0.9} &          92.9\textsubscript{\textpm1.6} &          98.1\textsubscript{\textpm0.8} &          60.7\textsubscript{\textpm0.8} &          86.1\textsubscript{\textpm0.4} \\
\bottomrule
\end{tabular}
\end{table*}

\section{Experiments}
In this section, we describe the details of a training procedure and present the outcomes from the experiments.

\subsection{Training Setup}
The PLM fine-tuning duration took maximally $100$ epochs with an early stopping mechanism~\cite{6120320} applied (patience $=10$ epochs). 
By using this technique, we were able to conserve computational resources while still obtaining high-performing models. 
The learning rate hyperparameter was always set to $2e-5$.
To prevent overfitting and enhance the model's generalization capabilities, we used a weight decay $w_d=0.01$ with Adam optimizer~\cite{8624183}.
The best performing model was selected based on F1-score achieved on the validation set, and the final results were reported on the test set (see Appendix~\ref{sec:training-details}). 
To minimize the influence of randomness on the outcomes, we trained PLM five times for each task using different initial seeds. 

During each experiment, the PLM was fine-tuned on ID data, which consisted of training and validation splits. 
The evaluation of the OOD detection methods themselves was performed on pre-defined test data.
A complete overview of the split sizes along with the number of classes in all data collections is presented in Table~\ref{tab:datasets}.

\subsection{Results}

\begin{table*}[ht]
    \centering
    \caption{AUROC (\%) and standard deviations for methods evaluated on datasets from second scenario. The first part of the table refers to semantic shift, where the second part refers to background shift.}
    \label{tab:scenario2}
    
\scalebox{0.89}{
\begin{tabular}{llllclllll}
\toprule
      \textbf{ID} &      \textbf{OOD} &                            \textbf{MSP} &                         \textbf{Energy} &                            \textbf{GradNorm} &                            \textbf{KLM} &                          \textbf{ReAct} &                            \textbf{DICE} &                                     \textbf{KNN} &                                     \textbf{ViM} \\
\midrule
   \multirow{2}{*}{Computer} & Politics & 91.5\textsubscript{\textpm1.9} & 96.3\textsubscript{\textpm1.1} & 95.5\textsubscript{\textpm0.9} & 78.0\textsubscript{\textpm7.3} & 96.2\textsubscript{\textpm1.2} & 34.6\textsubscript{\textpm13.2} &          97.0\textsubscript{\textpm0.5} & \textbf{98.6\textsubscript{\textpm0.3}} \\
 &   Sports & 89.8\textsubscript{\textpm2.7} & 94.9\textsubscript{\textpm1.6} & 94.1\textsubscript{\textpm1.6} & 74.5\textsubscript{\textpm4.6} & 94.6\textsubscript{\textpm1.7} &  51.9\textsubscript{\textpm6.9} &          95.7\textsubscript{\textpm0.9} & \textbf{97.7\textsubscript{\textpm0.6}} \\
\multirow{2}{*}{Politics} &    Computer & 94.4\textsubscript{\textpm0.8} & 96.0\textsubscript{\textpm0.6} & 95.5\textsubscript{\textpm0.7} & 82.8\textsubscript{\textpm4.6} & 95.9\textsubscript{\textpm0.6} &  63.9\textsubscript{\textpm3.2} &          96.9\textsubscript{\textpm0.2} & \textbf{98.3\textsubscript{\textpm0.2}} \\
 &   Sports & 91.4\textsubscript{\textpm1.1} & 93.4\textsubscript{\textpm0.9} & 92.9\textsubscript{\textpm1.0} & 72.3\textsubscript{\textpm5.6} & 93.3\textsubscript{\textpm0.9} &  58.6\textsubscript{\textpm2.4} &          95.3\textsubscript{\textpm0.4} & \textbf{97.3\textsubscript{\textpm0.3}} \\
  \multirow{2}{*}{Sports} &    Computer & 95.7\textsubscript{\textpm0.6} & 97.0\textsubscript{\textpm0.9} & 96.8\textsubscript{\textpm0.5} & 81.6\textsubscript{\textpm3.9} & 96.9\textsubscript{\textpm0.9} &  58.1\textsubscript{\textpm7.6} &          97.6\textsubscript{\textpm0.4} & \textbf{98.5\textsubscript{\textpm0.2}} \\
   & Politics & 95.3\textsubscript{\textpm0.2} & 96.5\textsubscript{\textpm0.6} & 96.4\textsubscript{\textpm0.5} & 79.9\textsubscript{\textpm2.5} & 96.5\textsubscript{\textpm0.7} & 52.4\textsubscript{\textpm11.5} &          97.2\textsubscript{\textpm0.3} & \textbf{98.0\textsubscript{\textpm0.1}} \\
  \midrule
    \multirow{2}{*}{IMDB} &    SST-2 & 85.3\textsubscript{\textpm0.8} & 84.3\textsubscript{\textpm1.8} & 77.8\textsubscript{\textpm3.0} & 61.2\textsubscript{\textpm1.7} & 84.5\textsubscript{\textpm1.9} &  84.6\textsubscript{\textpm3.3} & \textbf{97.8\textsubscript{\textpm1.2}} &          97.3\textsubscript{\textpm0.7} \\
    &     Yelp & 76.0\textsubscript{\textpm3.3} & 74.9\textsubscript{\textpm4.1} & 66.2\textsubscript{\textpm3.6} & 32.0\textsubscript{\textpm1.0} & 75.3\textsubscript{\textpm4.3} &  49.6\textsubscript{\textpm8.6} &          97.5\textsubscript{\textpm1.1} & \textbf{98.4\textsubscript{\textpm0.8}} \\
   \multirow{2}{*}{SST-2} &     IMDB & 83.2\textsubscript{\textpm1.4} & 82.7\textsubscript{\textpm2.2} & 70.3\textsubscript{\textpm2.3} & 55.0\textsubscript{\textpm2.7} & 83.3\textsubscript{\textpm2.4} & 34.5\textsubscript{\textpm10.7} & \textbf{87.2\textsubscript{\textpm1.7}} &          83.9\textsubscript{\textpm3.3} \\
    &     Yelp & 75.7\textsubscript{\textpm2.2} & 75.0\textsubscript{\textpm3.1} & 61.3\textsubscript{\textpm2.7} & 51.3\textsubscript{\textpm3.0} & 75.7\textsubscript{\textpm3.4} &  35.4\textsubscript{\textpm8.4} & \textbf{87.8\textsubscript{\textpm0.4}} &          80.1\textsubscript{\textpm2.8} \\
    \multirow{2}{*}{Yelp} &     IMDB & 79.5\textsubscript{\textpm0.5} & 79.2\textsubscript{\textpm1.6} & 71.7\textsubscript{\textpm1.9} & 38.6\textsubscript{\textpm1.3} & 79.5\textsubscript{\textpm1.6} &  26.8\textsubscript{\textpm5.1} &          84.7\textsubscript{\textpm0.8} & \textbf{88.6\textsubscript{\textpm0.7}} \\
     &    SST-2 & 91.6\textsubscript{\textpm0.5} & 91.5\textsubscript{\textpm0.9} & 86.1\textsubscript{\textpm1.0} & 59.9\textsubscript{\textpm2.5} & 91.7\textsubscript{\textpm0.9} &  55.8\textsubscript{\textpm8.5} &          98.5\textsubscript{\textpm0.3} & \textbf{99.0\textsubscript{\textpm0.1}} \\
\bottomrule
\end{tabular}
}
\end{table*} 

The outcomes from experiments on data prepared in the first scenario (Section~\ref{subsec:data_scenario_1}) are shown in Table~\ref{tab:scenario1}. 
The \textit{KNN} clearly outperformed the other OOD detection techniques on all three data groups. \textit{Energy-based} method also stands out with its good results as well as \textit{ViM}, except with its results on IMDB and Yelp dataset (worse than baseline \textit{MSP}). 
As expected, the values of evaluation metrics on the NC/O dataset were the lowest among \textit{Near-OOD} and \textit{Far-OOD} divisions. 
This dataset was separated from the original dataset used in the training, making it the most difficult to properly identify as OOD due to the distributional closeness.
The most challenging among the \textit{Far-OOD} collections appeared to be \textit{SST-2}, probably because of a small average number of words per example. 
The \textit{Language} turned out to be the easiest dataset to detect OOD samples, and almost all methods performed well on it.
The two worst performing approaches on the presented NLP tasks can be distinguished, i.e. \textit{DICE} and \textit{KLM}. 
Their measures were always worse than \textit{MSP}, sometimes even nearly random (a little above 50\%) -- \textit{DICE} on NC/O and \textit{KLM} on Twitter.

Interesting results can be seen in the last part of Table~\ref{tab:scenario1}. 
Randomization of words in case of NC/O dataset (which created NCR/O) significantly increased the model confidence in detecting OOD examples comparing with initial NC/O samples.
However, the OOD methods could not cope well with shuffled in-domain \textit{News category} data (NCR/I), which a human would recognize as the OOD.

\begin{figure}[ht]
    \centering
    \includegraphics[width=1.0\linewidth]{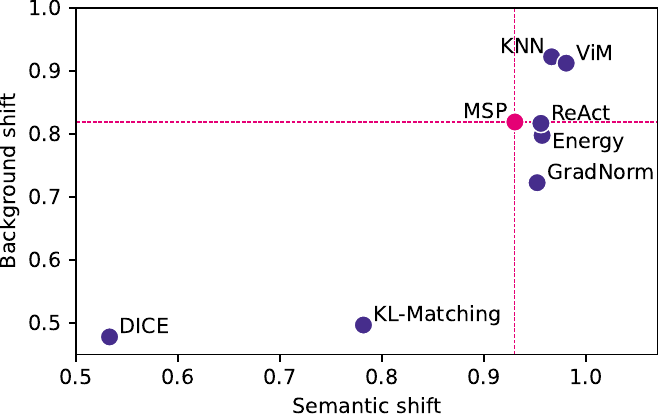}
    \caption{The performance of the methods is presented in AUROC depending on the type of distribution shift. The baseline method and its asymptotes are highlighted in pink color to facilitate comparison with other methods.}
    \label{fig:shift}
\end{figure}

Table~\ref{tab:scenario2} presents AUROC scores obtained from the second scenario (Section~\ref{subsec:data_scenario_2}) evaluation.
The results demonstrate that the \textit{ViM} method is more effective in detecting OOD samples with semantic shift to ID data. 
However, for background shift data, \textit{ViM} is not always the best and is outperformed by \textit{KNN} on IMDB and Yelp datasets. 
The SST-2 dataset proved to be problematic again, but only when used as a training set. 
It is worth noting that the average length of texts per SST-2 is considerably different from IMDB and Yelp collections, which mainly contain longer texts.
These observations suggest that \textit{KNN} is more stable in terms of different data characteristics.
To further emphasize the importance of comparing methods based on the type of shift, we created a visualization in Figure~\ref{fig:shift}.
The \textit{ReAct}, \textit{Energy}, and \textit{GradNorm} techniques turned out to be better than the baseline, but only for the semantic shift case.

To summarize, either \textit{KNN} or \textit{ViM} is the preferred choice among all the analyzed OOD detection approaches. Other reported metric values (AUPR-IN and FPR@95) from all experiments are attached in Appendix~\ref{sec:evaluation-details}. 

\subsection{Computational Resources}
All experiments were conducted on a workstation equipped with a mid-range \textit{Nvidia RTX 3060} GPU with 12GB of memory, a high-end \textit{Intel(R) Core(TM) i9-10900X} CPU with 20 cores and 40 threads, and 256 GB RAM.
These resources provided sufficient capacity for running the experiments and training the models used in this work, including analysis and processing of large datasets.
In total, we trained 35 models, taking 222 GPU-hours while evaluation alone lasted 124 GPU-hours.

\section{Conclusions}

The latest advancements in OOD detection techniques have surpassed the conventional \textit{MSP} baseline. 
In this work, we applied some of them to the NLP classification problems, selecting only post-hoc approaches because of their easy integration to already trained PLM model.
Most of the examined techniques achieved better results than the \textit{MSP}, but their performance varied when subjected to different types of data distributional shift.
Background shift was particularly challenging for the majority of methods to properly distinguish OOD examples. 
The \textit{KNN} and \textit{ViM} methods were found to be the most effective, and their performance was also stable. 
Hence, they are better alternatives to \textit{MSP} for out-of-distribution detection.
However, it should be kept in mind that it is likely that the \textit{ViM} method is sensitive to cases where the language model was trained on short texts and later exposed to a long text from outside the domain.

The proposed by us the unique analysis of \textit{Distinct-OOD} scenario, allowed to draw interesting findings. 
The tested methods were able to identify texts in different languages very easily as a OOD examples, but they had problems detecting OOD on the \textit{News Category Random} with shuffled data. 
This means that PLM models, despite their ability to detect contextual nuances in text, still tends to behave like Bag-of-Words~\cite{bag-of-words} in text classification tasks.
Business-wise, such structurally disturbed examples should not be further processed by AI systems. 
Therefore, OOD methods employed in NLP should better address semantic disorders in input sentences.

In conclusion, the overall performance of current OOD detection techniques is still low and unsatisfactory, particularly when presented with the \textit{Near-OOD} samples. 
Further research is necessary for the development of OOD detection methods, especially in the field of NLP, where more and more document processing automation systems are being developed, where ensuring reliability is important for users. 
Our work addresses the need for a comprehensive framework to evaluate the quality of OOD detection and provides easy extensibility to emerging methods.

\section{Limitations}

While our study provides valuable insights, it is important to keep in mind its limitations. Firstly, it was confined to text classification and did not include other NLP problems such as Named Entity Recognition (NER)~\cite{10.1145/3522593}, Question Answering (QA)~\cite{pandya2021question}, etc.
Expanding this research to a wider range of tasks would provide a better understanding of the methods' performance in diverse data scenarios. 
Additionally, the inclusion of a task shift can be valuable, where the model is trained on a single task but OOD data come from a totally different prediction problems.

Secondly, we conducted our experiments using only RoBERTa model. 
We chose a widely used language model for text classification, but there are several other architectures worth testing, especially large language models (LLMs)~\cite{DBLP:journals/corr/abs-2303-18223} that now becoming extremely popular.
A more comprehensive evaluation of the models and methods could provide more insights into whether the development of transformer-based methods contributes to better detection of OOD~data.

Finally, due to restricted computational time, we did not perform a hyperparameter search for either model or methods. 
We just used recommend values from the original publications.
This may have affected the obtained results, and it is certainly an aspect worth investigating in the future.

\section{Ethics Statement}

The authors believe that their work does not raise any ethical questions of harm or discrimination. Moreover, they acknowledge that the benchmark has a wide range of potential applications and want to make it clear that they are not responsible for any unethical applications of their work.

\section*{Acknowledgements}

The research was conducted under the Implementation Doctorate programme of Polish Ministry of Science and Higher Education (DWD/6/0322/2022) with cooperation of the Artificial Intelligence Department at Wroclaw University of Science and Technology.
It was partially co-funded by the European Regional Development Fund within the Priority Axis 1 “Enterprises and innovation”, Measure 1.2. “Innovative enterprises, sub-measure 1.2.1. “Innovative enterprises – horizontal competition” as part of ROP WD 2014-2020, support contract no. RPDS.01.02.01-02-0063/20-00.
The work conducted by Maciej Zieba was supported by the National Centre of Science (Poland) Grant No. 2021/43/B/ST6/02853.

\bibliography{anthology,custom}
\bibliographystyle{acl_natbib}

\appendix

\section{Training Details}
\label{sec:training-details}

Each model was trained on five different seeds from range $[2021, 2025]$.
Table~\ref{tab:training-metrics} includes averaged classification metrics with standard deviation.

\begin{table}[ht]
    \centering
    \caption{Training metrics on test set.}
    \label{tab:training-metrics}
    
\scalebox{0.82}{
\begin{tabular}{lcccc}
\toprule
 \textbf{Dataset} &                       \textbf{Accuracy} &                             \textbf{F1 Score} &                      \textbf{Precision} &                         \textbf{Recall} \\
\midrule
    NC/I & 82.4\textsubscript{\textpm0.1} & 81.8\textsubscript{\textpm0.1} & 81.7\textsubscript{\textpm0.2} & 82.0\textsubscript{\textpm0.2} \\
Computer & 89.2\textsubscript{\textpm0.3} & 89.3\textsubscript{\textpm0.4} & 89.3\textsubscript{\textpm0.4} & 89.3\textsubscript{\textpm0.3} \\
Politics & 94.7\textsubscript{\textpm0.3} & 94.6\textsubscript{\textpm0.3} & 94.6\textsubscript{\textpm0.4} & 94.7\textsubscript{\textpm0.3} \\
  Sports & 97.5\textsubscript{\textpm0.2} & 97.5\textsubscript{\textpm0.2} & 97.5\textsubscript{\textpm0.2} & 97.5\textsubscript{\textpm0.2} \\
    IMDB & 94.7\textsubscript{\textpm0.1} & 94.7\textsubscript{\textpm0.1} & 94.7\textsubscript{\textpm0.1} & 94.7\textsubscript{\textpm0.1} \\
   SST-2 & 93.9\textsubscript{\textpm0.1} & 93.8\textsubscript{\textpm0.1} & 93.7\textsubscript{\textpm0.1} & 93.8\textsubscript{\textpm0.1} \\
    Yelp & 96.9\textsubscript{\textpm0.0} & 96.9\textsubscript{\textpm0.0} & 96.9\textsubscript{\textpm0.0} & 96.9\textsubscript{\textpm0.0} \\
\bottomrule
\end{tabular}
}
\end{table} 

\section{Evaluation Details}
\label{sec:evaluation-details}

The values for all metrics that were considered in our experiments are listed below.
Tables \ref{tab:evaluation-scenario1-auprin} and \ref{tab:evaluation-scenario1-fpr} refer to the Scenario~1 of OOD data preparation; Tables \ref{tab:evaluation-scenario2-auprin} and \ref{tab:evaluation-scenario2-fpr} report the results from the Scenario~2.

\begin{table*}[ht]
    \centering
    \caption{AUPR-IN (\%) and standard deviations for methods evaluated on datasets from first scenario.}
    \label{tab:evaluation-scenario1-auprin}
    
\scalebox{0.88}{
\begin{tabular}{lllllllll}
\toprule
\textbf{Method} &                           \textbf{NC/O} &                        \textbf{Twitter} &                           \textbf{IMDB} &                          \textbf{SST-2} &                           \textbf{Yelp} &                       \textbf{Language} &                          \textbf{NCR/I} &                          \textbf{NCR/O} \\
\midrule
            MSP &          71.7\textsubscript{\textpm0.4} &          97.3\textsubscript{\textpm0.3} &          98.4\textsubscript{\textpm1.5} &          91.9\textsubscript{\textpm1.8} &          97.4\textsubscript{\textpm0.8} &          99.2\textsubscript{\textpm0.3} &          59.2\textsubscript{\textpm1.1} &          80.4\textsubscript{\textpm0.7} \\
         Energy &          74.5\textsubscript{\textpm0.6} &          98.5\textsubscript{\textpm0.2} &          99.8\textsubscript{\textpm0.2} &          96.3\textsubscript{\textpm1.3} &          99.2\textsubscript{\textpm0.4} &          99.8\textsubscript{\textpm0.1} &          58.8\textsubscript{\textpm1.3} &          84.0\textsubscript{\textpm0.9} \\
       GradNorm &          73.9\textsubscript{\textpm0.7} &          98.2\textsubscript{\textpm0.3} &          99.5\textsubscript{\textpm0.6} &          95.4\textsubscript{\textpm1.1} &          98.8\textsubscript{\textpm0.4} &          99.7\textsubscript{\textpm0.1} &          58.5\textsubscript{\textpm1.9} &          83.8\textsubscript{\textpm1.0} \\
            KL-Matching &          51.0\textsubscript{\textpm0.4} &          90.9\textsubscript{\textpm0.7} &          94.1\textsubscript{\textpm5.1} &          72.0\textsubscript{\textpm2.8} &          87.7\textsubscript{\textpm3.6} &          96.6\textsubscript{\textpm1.3} &          48.3\textsubscript{\textpm0.2} &          54.8\textsubscript{\textpm0.5} \\
          ReAct &          74.3\textsubscript{\textpm0.5} &          98.4\textsubscript{\textpm0.2} &          99.8\textsubscript{\textpm0.2} &          96.1\textsubscript{\textpm1.4} &          99.2\textsubscript{\textpm0.4} &          99.8\textsubscript{\textpm0.1} &          58.9\textsubscript{\textpm1.4} &          83.7\textsubscript{\textpm0.9} \\
           DICE &          51.8\textsubscript{\textpm0.7} &          96.0\textsubscript{\textpm0.4} &          91.0\textsubscript{\textpm2.6} &          82.6\textsubscript{\textpm0.9} &          94.1\textsubscript{\textpm0.9} &          95.0\textsubscript{\textpm0.5} &          51.0\textsubscript{\textpm1.0} &          56.7\textsubscript{\textpm1.5} \\
            KNN & \textbf{78.5\textsubscript{\textpm0.1}} & \textbf{99.3\textsubscript{\textpm0.1}} & \textbf{99.9\textsubscript{\textpm0.0}} & \textbf{98.3\textsubscript{\textpm0.5}} & \textbf{99.8\textsubscript{\textpm0.1}} & \textbf{99.9\textsubscript{\textpm0.0}} & \textbf{68.9\textsubscript{\textpm1.3}} & \textbf{88.6\textsubscript{\textpm0.6}} \\
            VIM &          77.1\textsubscript{\textpm0.1} &          99.0\textsubscript{\textpm0.2} &          96.5\textsubscript{\textpm1.1} &          98.1\textsubscript{\textpm0.5} &          97.1\textsubscript{\textpm0.6} &          99.7\textsubscript{\textpm0.1} &          58.9\textsubscript{\textpm0.9} &          85.5\textsubscript{\textpm0.4} \\
\bottomrule
\end{tabular}
}
\end{table*} 

\begin{table*}[ht]
    \centering
    \caption{FPR@95 (\%) and standard deviations for methods evaluated on datasets from first scenario. Lower scores indicate better performance.}
    \label{tab:evaluation-scenario1-fpr}
    
\scalebox{0.85}{
    \begin{tabular}{lllllllll}
\toprule
\textbf{Method} &                           \textbf{NC/O} &                        \textbf{Twitter} &                          \textbf{IMDB} &                          \textbf{SST-2} &                          \textbf{Yelp} &                      \textbf{Language} &                          \textbf{NCR/I} &                          \textbf{NCR/O} \\
\midrule
            MSP &          82.3\textsubscript{\textpm0.8} &          77.3\textsubscript{\textpm4.8} &        19.6\textsubscript{\textpm18.8} &          61.3\textsubscript{\textpm7.8} &         21.5\textsubscript{\textpm6.7} &        29.3\textsubscript{\textpm10.7} &          91.3\textsubscript{\textpm0.5} &          75.2\textsubscript{\textpm1.4} \\
         Energy &          75.2\textsubscript{\textpm1.0} &          55.7\textsubscript{\textpm7.3} &          2.4\textsubscript{\textpm2.7} &         35.8\textsubscript{\textpm10.5} &          7.1\textsubscript{\textpm3.5} &          7.6\textsubscript{\textpm3.5} &          89.0\textsubscript{\textpm0.6} &          63.8\textsubscript{\textpm1.9} \\
       GradNorm &          75.9\textsubscript{\textpm0.8} &          65.1\textsubscript{\textpm6.9} &          5.7\textsubscript{\textpm6.3} &          44.0\textsubscript{\textpm7.9} &         11.2\textsubscript{\textpm4.0} &         12.9\textsubscript{\textpm5.0} &          88.8\textsubscript{\textpm0.7} &          63.7\textsubscript{\textpm2.1} \\
            KL-Matching &          85.8\textsubscript{\textpm0.5} &          85.4\textsubscript{\textpm3.5} &        33.8\textsubscript{\textpm29.7} &          76.4\textsubscript{\textpm4.6} &         30.2\textsubscript{\textpm8.7} &         55.7\textsubscript{\textpm9.2} &          92.3\textsubscript{\textpm0.3} &          80.2\textsubscript{\textpm0.8} \\
          ReAct &          75.3\textsubscript{\textpm1.1} &          55.3\textsubscript{\textpm7.2} &          2.2\textsubscript{\textpm2.5} &         35.6\textsubscript{\textpm10.6} &          7.0\textsubscript{\textpm3.4} &          7.6\textsubscript{\textpm3.6} &          89.2\textsubscript{\textpm0.6} &          64.2\textsubscript{\textpm1.9} \\
           DICE &          95.2\textsubscript{\textpm0.3} &          99.9\textsubscript{\textpm0.0} &        100.0\textsubscript{\textpm0.0} &          99.9\textsubscript{\textpm0.1} &         99.4\textsubscript{\textpm0.9} &        100.0\textsubscript{\textpm0.0} &          96.3\textsubscript{\textpm0.3} &          97.1\textsubscript{\textpm0.5} \\
            KNN &          73.9\textsubscript{\textpm0.6} & \textbf{34.4\textsubscript{\textpm5.4}} & \textbf{0.2\textsubscript{\textpm0.1}} & \textbf{22.1\textsubscript{\textpm8.6}} & \textbf{2.2\textsubscript{\textpm0.7}} & \textbf{1.4\textsubscript{\textpm0.5}} & \textbf{85.7\textsubscript{\textpm0.7}} & \textbf{56.1\textsubscript{\textpm1.6}} \\
            VIM & \textbf{71.5\textsubscript{\textpm0.5}} &          57.8\textsubscript{\textpm4.7} &        86.5\textsubscript{\textpm12.4} &          23.7\textsubscript{\textpm5.2} &        63.3\textsubscript{\textpm10.7} &         13.2\textsubscript{\textpm8.0} &          88.9\textsubscript{\textpm0.5} &          63.4\textsubscript{\textpm1.1} \\
\bottomrule
\end{tabular}
}
\end{table*} 

\begin{table*}[ht]
    \centering
    \caption{AUPR-IN (\%) and standard deviations for methods evaluated on datasets from second scenario. The first part of the table refers to semantic shift, where the second part refers to background shift.}
    \label{tab:evaluation-scenario2-auprin}
    
\scalebox{0.80}{
\begin{tabular}{llllllllll}
\toprule
\textbf{ID} & \textbf{OOD} &                   \textbf{MSP} &                \textbf{Energy} &                   \textbf{GradNorm} &                   \textbf{KLM} &                 \textbf{ReAct} &                   \textbf{DICE} &                            \textbf{KNN} &                            \textbf{VIM} \\
\midrule
   \multirow{2}{*}{Computer} &     Politics & 95.2\textsubscript{\textpm1.1} & 97.7\textsubscript{\textpm0.7} & 97.4\textsubscript{\textpm0.5} & 77.7\textsubscript{\textpm8.2} & 97.6\textsubscript{\textpm0.7} & 56.1\textsubscript{\textpm11.4} &          98.2\textsubscript{\textpm0.3} & \textbf{99.1\textsubscript{\textpm0.2}} \\
    &       Sports & 93.3\textsubscript{\textpm1.9} & 96.4\textsubscript{\textpm1.1} & 96.0\textsubscript{\textpm1.0} & 71.3\textsubscript{\textpm5.5} & 96.2\textsubscript{\textpm1.2} &  64.3\textsubscript{\textpm9.0} &          97.1\textsubscript{\textpm0.6} & \textbf{98.3\textsubscript{\textpm0.4}} \\
   \multirow{2}{*}{Politics} &     Computer & 93.8\textsubscript{\textpm0.7} & 94.8\textsubscript{\textpm0.7} & 94.6\textsubscript{\textpm0.7} & 67.3\textsubscript{\textpm9.0} & 94.7\textsubscript{\textpm0.7} &  68.9\textsubscript{\textpm2.2} &          96.7\textsubscript{\textpm0.2} & \textbf{97.9\textsubscript{\textpm0.2}} \\
    &       Sports & 91.6\textsubscript{\textpm1.2} & 92.8\textsubscript{\textpm1.0} & 92.4\textsubscript{\textpm1.2} & 60.8\textsubscript{\textpm9.6} & 92.6\textsubscript{\textpm1.1} &  67.5\textsubscript{\textpm1.9} &          95.8\textsubscript{\textpm0.3} & \textbf{97.1\textsubscript{\textpm0.3}} \\
     \multirow{2}{*}{Sports} &     Computer & 96.3\textsubscript{\textpm0.7} & 96.9\textsubscript{\textpm1.1} & 97.1\textsubscript{\textpm0.5} & 70.1\textsubscript{\textpm7.4} & 96.8\textsubscript{\textpm1.1} &  67.2\textsubscript{\textpm6.4} &          98.0\textsubscript{\textpm0.3} & \textbf{98.7\textsubscript{\textpm0.2}} \\
      &     Politics & 96.6\textsubscript{\textpm0.4} & 97.1\textsubscript{\textpm0.9} & 97.4\textsubscript{\textpm0.5} & 75.3\textsubscript{\textpm1.7} & 97.1\textsubscript{\textpm0.9} &  66.0\textsubscript{\textpm9.7} &          98.2\textsubscript{\textpm0.2} & \textbf{98.6\textsubscript{\textpm0.1}} \\
      \midrule
       \multirow{2}{*}{IMDB} &        SST-2 & 86.2\textsubscript{\textpm1.4} & 84.8\textsubscript{\textpm1.8} & 73.7\textsubscript{\textpm6.6} & 52.2\textsubscript{\textpm1.2} & 85.0\textsubscript{\textpm1.8} &  85.5\textsubscript{\textpm3.6} & \textbf{98.1\textsubscript{\textpm1.0}} &          97.6\textsubscript{\textpm0.6} \\
        &         Yelp & 82.1\textsubscript{\textpm2.8} & 80.8\textsubscript{\textpm3.6} & 71.5\textsubscript{\textpm3.3} & 38.8\textsubscript{\textpm0.6} & 81.2\textsubscript{\textpm3.9} &  51.4\textsubscript{\textpm8.0} &          97.9\textsubscript{\textpm0.8} & \textbf{98.6\textsubscript{\textpm0.6}} \\
      \multirow{2}{*}{SST-2} &         IMDB & 85.7\textsubscript{\textpm1.5} & 85.1\textsubscript{\textpm2.0} & 69.4\textsubscript{\textpm3.0} & 48.6\textsubscript{\textpm1.4} & 85.7\textsubscript{\textpm2.2} &  41.1\textsubscript{\textpm5.0} & \textbf{91.4\textsubscript{\textpm0.8}} &          86.5\textsubscript{\textpm2.5} \\
       &         Yelp & 76.3\textsubscript{\textpm2.8} & 75.4\textsubscript{\textpm3.5} & 60.5\textsubscript{\textpm3.5} & 47.4\textsubscript{\textpm1.5} & 76.3\textsubscript{\textpm3.8} &  40.6\textsubscript{\textpm3.6} & \textbf{91.4\textsubscript{\textpm0.4}} &          82.5\textsubscript{\textpm2.6} \\
       \multirow{2}{*}{Yelp} &         IMDB & 83.5\textsubscript{\textpm0.5} & 82.7\textsubscript{\textpm2.5} & 76.1\textsubscript{\textpm2.3} & 41.1\textsubscript{\textpm0.5} & 83.0\textsubscript{\textpm2.4} &  36.8\textsubscript{\textpm1.6} &          88.2\textsubscript{\textpm0.6} & \textbf{91.2\textsubscript{\textpm0.5}} \\
        &        SST-2 & 93.8\textsubscript{\textpm0.4} & 93.7\textsubscript{\textpm0.7} & 88.8\textsubscript{\textpm0.8} & 50.2\textsubscript{\textpm1.6} & 93.9\textsubscript{\textpm0.7} &  63.3\textsubscript{\textpm8.2} &          98.9\textsubscript{\textpm0.2} & \textbf{99.3\textsubscript{\textpm0.1}} \\
\bottomrule
\end{tabular}
}
\end{table*} 

\begin{table*}[ht]
    \centering
    \caption{FPR@95 (\%) and standard deviations for methods evaluated on datasets from second scenario. The first part of the table refers to semantic shift, where the second part refers to background shift. Lower scores indicate better performance.}
    \label{tab:evaluation-scenario2-fpr}
    
\scalebox{0.77}{
\begin{tabular}{llllllllll}
\toprule
\textbf{ID} & \textbf{OOD} &                            \textbf{MSP} &                \textbf{Energy} &                   \textbf{GradNorm} &                    \textbf{KLM} &                  \textbf{ReAct} &                   \textbf{DICE} &                            \textbf{KNN} &                            \textbf{VIM} \\
\midrule
   \multirow{2}{*}{Computer} &     Politics &         55.9\textsubscript{\textpm11.7} & 20.9\textsubscript{\textpm7.4} & 31.0\textsubscript{\textpm8.4} & 61.6\textsubscript{\textpm12.6} &  21.3\textsubscript{\textpm7.5} &  99.9\textsubscript{\textpm0.1} &          17.3\textsubscript{\textpm4.6} &  \textbf{7.2\textsubscript{\textpm1.8}} \\
    &       Sports &          61.4\textsubscript{\textpm9.3} & 30.8\textsubscript{\textpm8.8} & 39.7\textsubscript{\textpm8.7} &  66.7\textsubscript{\textpm9.6} &  31.4\textsubscript{\textpm8.6} &  99.1\textsubscript{\textpm0.9} &          29.6\textsubscript{\textpm6.3} & \textbf{14.1\textsubscript{\textpm5.6}} \\
   \multirow{2}{*}{Politics} &     Computer &          38.4\textsubscript{\textpm8.9} & 22.0\textsubscript{\textpm4.0} & 28.1\textsubscript{\textpm8.9} &  42.1\textsubscript{\textpm9.8} &  22.7\textsubscript{\textpm4.1} &  98.8\textsubscript{\textpm0.9} &          22.8\textsubscript{\textpm4.6} &  \textbf{9.4\textsubscript{\textpm1.6}} \\
    &       Sports &          55.8\textsubscript{\textpm7.4} & 35.6\textsubscript{\textpm4.7} & 42.5\textsubscript{\textpm9.4} &  59.4\textsubscript{\textpm7.8} &  36.5\textsubscript{\textpm4.7} &  99.4\textsubscript{\textpm0.5} &          35.8\textsubscript{\textpm5.0} & \textbf{16.2\textsubscript{\textpm3.1}} \\
     \multirow{2}{*}{Sports} &     Computer &          27.9\textsubscript{\textpm6.6} & 18.1\textsubscript{\textpm5.7} & 18.2\textsubscript{\textpm5.1} &  32.2\textsubscript{\textpm5.1} &  18.8\textsubscript{\textpm6.0} &  96.0\textsubscript{\textpm2.3} &          11.8\textsubscript{\textpm4.2} &  \textbf{6.0\textsubscript{\textpm1.6}} \\
      &     Politics &          30.5\textsubscript{\textpm3.6} & 21.0\textsubscript{\textpm2.8} & 21.0\textsubscript{\textpm2.9} &  33.9\textsubscript{\textpm2.3} &  21.7\textsubscript{\textpm3.3} &  95.5\textsubscript{\textpm7.4} &          17.7\textsubscript{\textpm2.6} &  \textbf{9.1\textsubscript{\textpm1.2}} \\
      \midrule
       \multirow{2}{*}{IMDB} &        SST-2 &          65.6\textsubscript{\textpm0.9} & 68.5\textsubscript{\textpm9.4} & 67.3\textsubscript{\textpm1.6} &  65.6\textsubscript{\textpm0.9} & 69.1\textsubscript{\textpm10.7} &  54.4\textsubscript{\textpm9.6} & \textbf{12.5\textsubscript{\textpm8.2}} &          14.0\textsubscript{\textpm3.9} \\
        &         Yelp &          92.3\textsubscript{\textpm1.3} & 92.8\textsubscript{\textpm2.8} & 93.3\textsubscript{\textpm1.0} &  92.3\textsubscript{\textpm1.3} &  92.6\textsubscript{\textpm2.9} &  93.8\textsubscript{\textpm7.2} &          15.1\textsubscript{\textpm9.1} &  \textbf{8.2\textsubscript{\textpm6.5}} \\
      \multirow{2}{*}{SST-2} &         IMDB & \textbf{77.7\textsubscript{\textpm2.3}} & 79.6\textsubscript{\textpm7.3} & 81.8\textsubscript{\textpm1.5} &  78.0\textsubscript{\textpm2.3} &  79.1\textsubscript{\textpm8.9} & 100.0\textsubscript{\textpm0.0} &         88.3\textsubscript{\textpm10.9} &         79.0\textsubscript{\textpm17.8} \\
       &         Yelp &          84.5\textsubscript{\textpm2.4} & 85.8\textsubscript{\textpm6.2} & 87.6\textsubscript{\textpm1.2} &  84.8\textsubscript{\textpm2.3} &  85.2\textsubscript{\textpm7.0} &  99.7\textsubscript{\textpm0.2} & \textbf{81.0\textsubscript{\textpm9.0}} &         82.2\textsubscript{\textpm13.1} \\
       \multirow{2}{*}{Yelp} &         IMDB &          83.7\textsubscript{\textpm0.6} & 83.5\textsubscript{\textpm1.4} & 87.1\textsubscript{\textpm0.7} &  83.6\textsubscript{\textpm0.6} &  83.2\textsubscript{\textpm1.4} &  99.7\textsubscript{\textpm0.1} &          74.9\textsubscript{\textpm3.8} & \textbf{62.4\textsubscript{\textpm2.5}} \\
        &        SST-2 &          58.4\textsubscript{\textpm2.4} & 58.6\textsubscript{\textpm4.0} & 66.7\textsubscript{\textpm2.5} &  58.4\textsubscript{\textpm2.4} &  57.9\textsubscript{\textpm4.2} &  96.1\textsubscript{\textpm4.7} &           3.8\textsubscript{\textpm1.5} &  \textbf{2.4\textsubscript{\textpm0.8}} \\
\bottomrule
\end{tabular}
}
\end{table*} 


\end{document}